\PassOptionsToPackage{table,xcdraw}{xcolor}
\documentclass[sigconf,natbib=true]{acmart}

\usepackage{booktabs}
\usepackage{amssymb}
\usepackage{amsmath} 
\usepackage{bm}
\usepackage{courier}
\usepackage{enumitem}
\usepackage{bbding}
\usepackage{pifont}
\usepackage{setspace}
\usepackage{array,multirow,multicol}
\usepackage{caption, subcaption}
\usepackage{algorithm} 
\usepackage{algpseudocode} 
\usepackage{lipsum}
\usepackage{etoolbox}

\usepackage{graphicx}

\newcommand{\PreserveBackslash}[1]{\let\temp=\\#1\let\\=\temp}
\newcolumntype{C}[1]{>{\PreserveBackslash\centering}p{#1}}
\newcolumntype{R}[1]{>{\PreserveBackslash\raggedleft}p{#1}}
\newcolumntype{L}[1]{>{\PreserveBackslash\raggedright}p{#1}}
  
\newtheorem{theorem}{Theorem}

\usepackage[table,xcdraw]{xcolor}

\DeclareMathOperator*{\argmin}{arg\,min}
\usepackage{tikz}

\AtBeginDocument{%
  \providecommand\BibTeX{{%
    \normalfont B\kern-0.5em{\scshape i\kern-0.25em b}\kern-0.8em\TeX}}}

\copyrightyear{2023} 
\acmYear{2023} 
\setcopyright{acmlicensed}\acmConference[SIGIR '23]{Proceedings of the 46th International ACM SIGIR Conference on Research and Development in Information Retrieval}{July 23--27, 2023}{Taipei, Taiwan}
\acmBooktitle{Proceedings of the 46th International ACM SIGIR Conference on Research and Development in Information Retrieval (SIGIR '23), July 23--27, 2023, Taipei, Taiwan}
\acmPrice{15.00}
\acmDOI{10.1145/3539618.3592059}
\acmISBN{978-1-4503-9408-6/23/07}

\begin{document}

\title{Sharpness-Aware Graph Collaborative Filtering}

\author{Huiyuan Chen}
\email{hchen@visa.com}
\affiliation{%
  \institution{Visa Research}
  \country{}
}

\author{Chin-Chia Michael Yeh}
\email{miyeh@visa.com}
\affiliation{%
  \institution{Visa Research}
  \country{}
}

\author{Yujie Fan}
\email{yufan@visa.com}
\affiliation{%
  \institution{Visa Research}
  \country{}
}

\author{Yan Zheng}
\email{yazheng@visa.com}
\affiliation{%
  \institution{Visa Research}
  \country{}
}

\author{Junpeng Wang}
\email{junpenwa@visa.com}
\affiliation{%
  \institution{Visa Research}
  \country{}
}

\author{Vivian Lai}
\email{viv.lai@visa.com}
\affiliation{%
  \institution{Visa Research}
  \country{}
}

\author{Mahashweta Das}
\email{mahdas@visa.com}
\affiliation{%
  \institution{Visa Research}
  \country{}
}

\author{Hao Yang}
\email{haoyang@visa.com}
\affiliation{%
  \institution{Visa Research}
  \country{}
}
\renewcommand{\shortauthors}{Huiyuan Chen et al.}
\begin{abstract}
Graph Neural Networks (GNNs) have achieved impressive performance  in  collaborative filtering.   However,   GNNs tend to yield inferior performance when the distributions of training and test data are not aligned well. Also,  training GNNs  requires optimizing   non-convex neural networks with an abundance of local and global minima, which may differ widely in their performance at test time.   Thus, it is essential to   choose the minima carefully. Here we propose an effective training schema, called {gSAM}, under the principle that the \textit{flatter}  minima has a better generalization ability than the \textit{sharper} ones.  To achieve this goal, gSAM  regularizes the flatness of the weight loss landscape by forming a bi-level optimization: the outer problem conducts the standard model training while the inner problem helps the model jump out of the sharp minima. Experimental results show the superiority of our gSAM.
\end{abstract}

\maketitle

\section{Introduction}
Collaborative Filtering (CF) has been widely used in recommender systems due to its efficiency~\cite{schafer2007collaborative,koren2009matrix,rendle2009bpr,covington2016deep}. 
As the user-item interaction data can be naturally represented as a bipartite graph, Graph Neural Networks (GNNs)~\cite{ying2018graph,wang2019neural,yeh2022embedding,he2020lightgcn,yu2022graph,wang2022improving,chen2021structured,chen2022tinykg} have gained considerable attention to fully leverage graph structural information. As such,  higher-order dependencies between users and items can be captured  to refine the node embeddings.

However, real-world graphs  often   exhibit a power-law distribution~\cite{hu2020open}, where the long tail contains low-degree   items that lack collaborative signals. Hence, the neighborhood
aggregation scheme in GNNs is inevitably biased towards high-degree items~\cite{wu2021self}, neglecting the impact of low-degree items. Also,   the training and the test data distribution may differ significantly due to distribution shifts~\cite{wang2020streaming}. This notoriously makes the  training of GNNs suffer from overfitting issues with poor generalization  and stability.

On the other hand, GNNs, inheriting the proprieties of neural networks, are often overparameterized and have the capacity to fit even a  random  labeling  of  the  training  data~\cite{zhang2021understanding}. Thus, a small training loss does not necessarily guarantee good generalization. Additionally,  training GNNs often requires optimizing complex and non-convex neural networks, with an abundance of local and global minima. Indeed, not all  minima are created equal,  and  several studies~\cite{neyshabur2017exploring,foret2020sharpness,liu2020bad,kawaguchi2016deep} show that the standard Stochastic Gradient Descent (SGD) can be easily made to converge to bad minima (e.g., saddle points or sharp minima) that could not generalize well.

\begin{figure}
\centering
\begin{minipage}{0.38\linewidth}
\centering
\includegraphics[width=0.72\linewidth,angle=90]{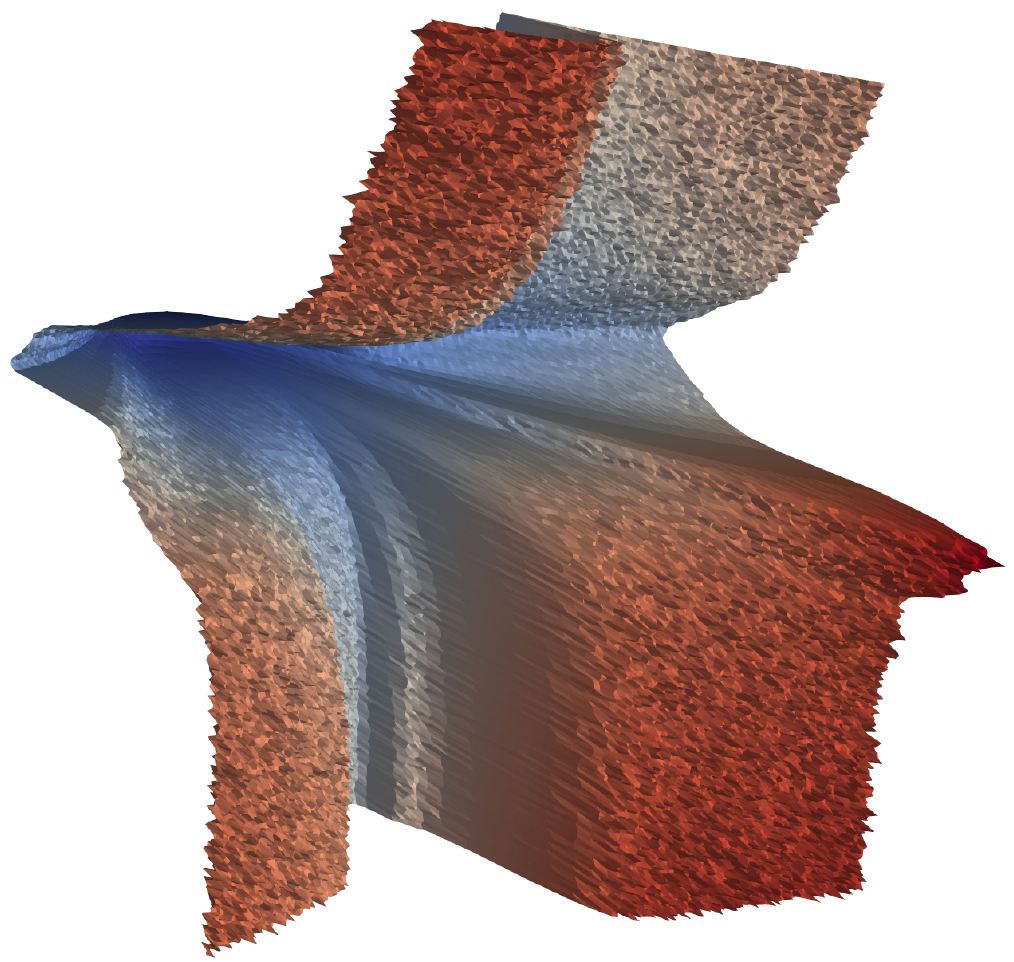}
\caption*{(a)  NGCF.}
\end{minipage}\hfill
\begin{minipage}{0.38\linewidth}
\centering
\includegraphics[width=0.7\linewidth,angle=-90]{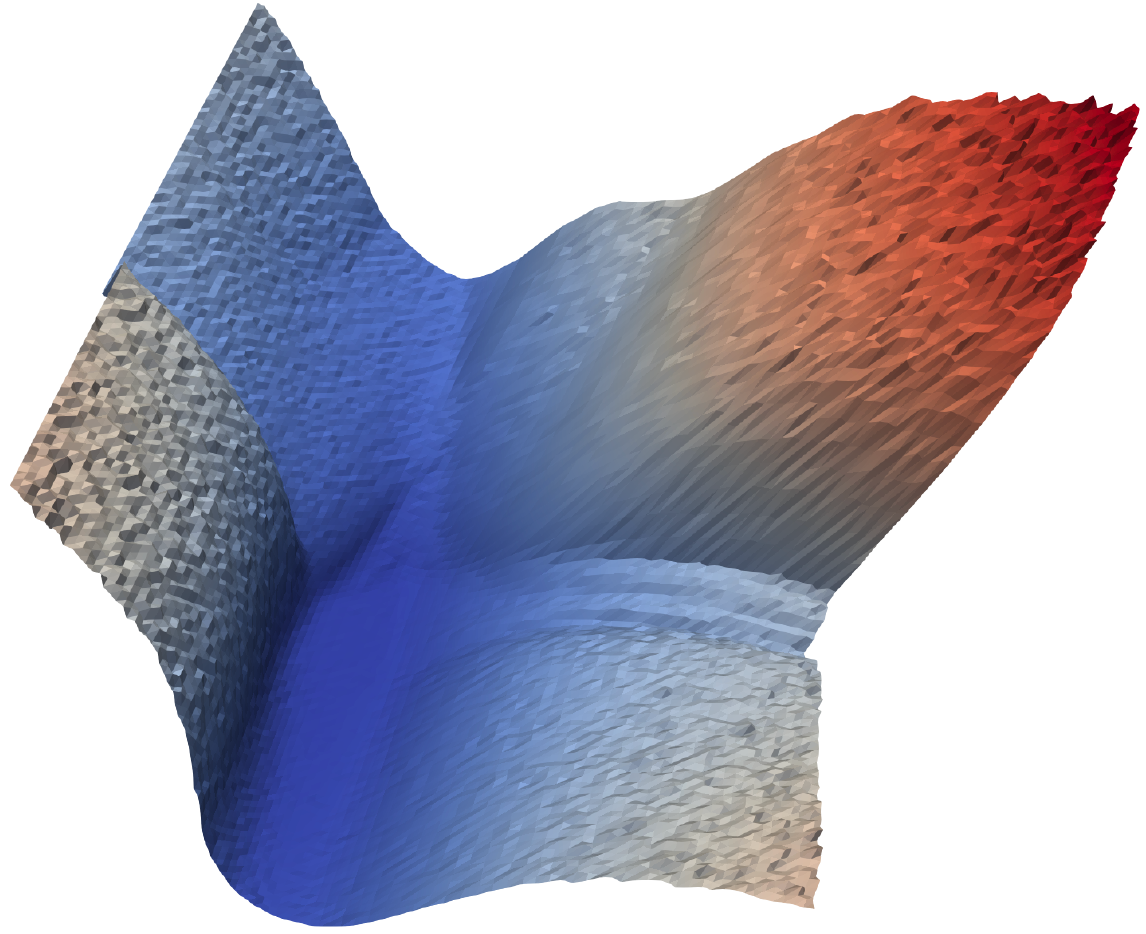}
\caption*{(b) NGCF+gSAM.}
\end{minipage}
\begin{minipage}[t]{0.38\linewidth}
\centering
\includegraphics[width=0.75\linewidth,angle=-90]{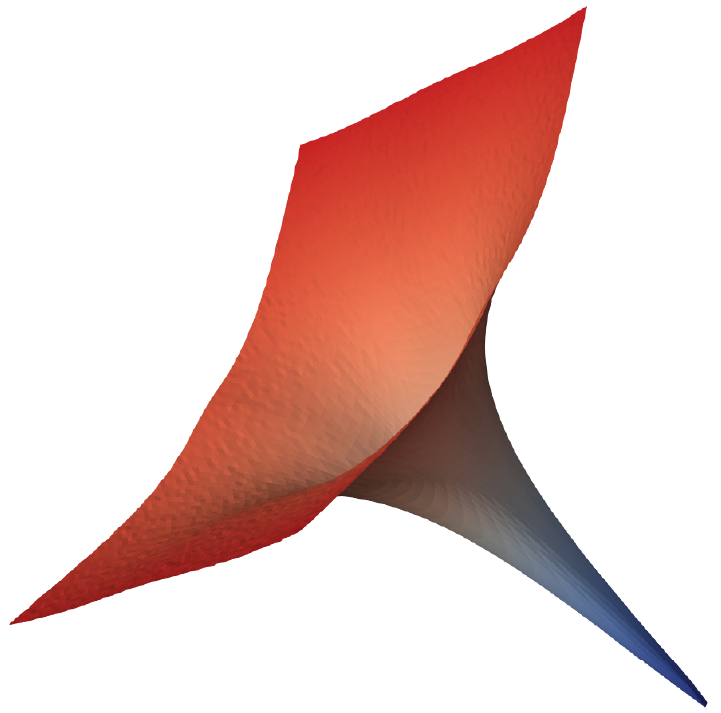}
\caption*{(c)  LightGCN.}
\end{minipage}\hfill
\begin{minipage}[t]{0.38\linewidth}
\centering
\includegraphics[width=0.75\linewidth,angle=-90]{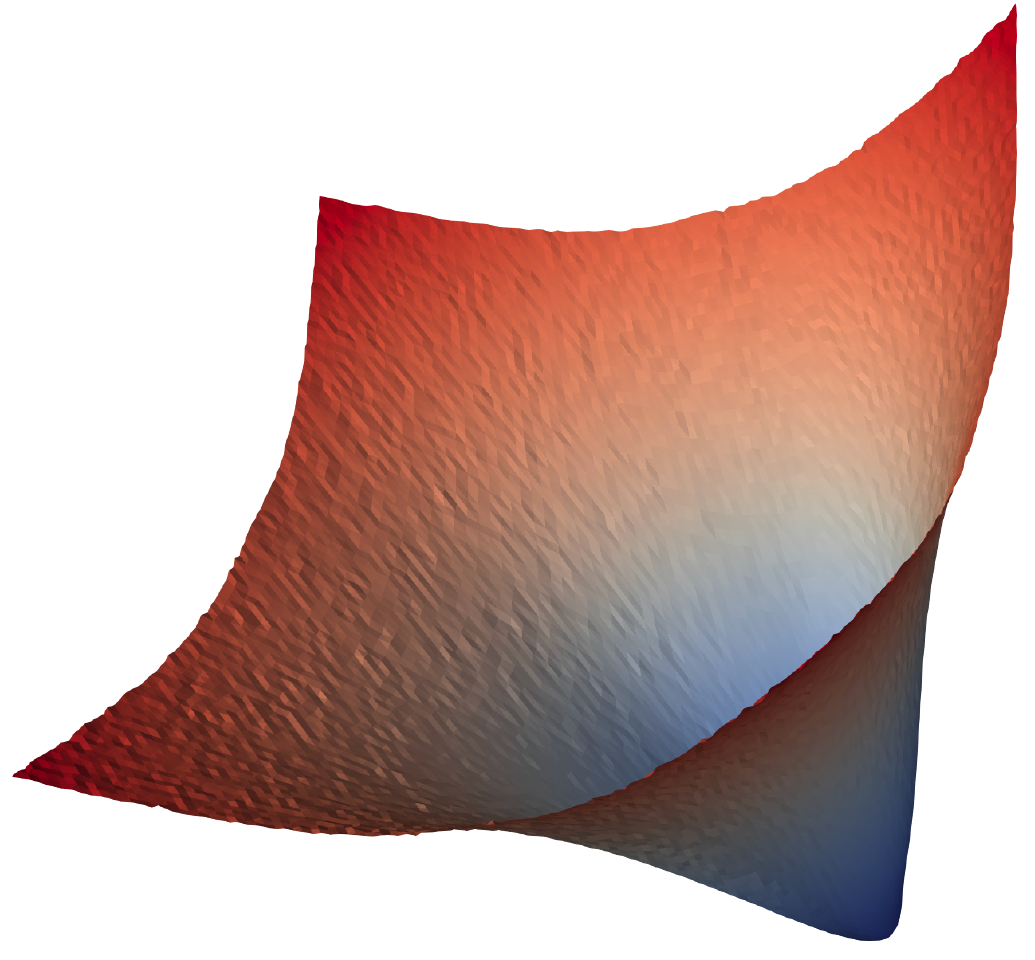}
\caption*{(d) LightGCN+gSAM.}
\end{minipage}
\caption{Compared to vanilla baselines (\textsl{e.g.}, NGCF and LightGCN) with sharp surfaces, our training schema gSAM produces a much smoother surface on the Amazon-Book dataset. }
\label{figure1}
\vspace{-1mm}
\end{figure}

To explain the above concerns, we   explore the  loss landscape of  NGCF and LightGCN ~\cite{li2018visualizing,foret2020sharpness}. Given a well-trained GNN model $f_{\boldsymbol{\theta}}$, we  compute the loss values when moving  the model parameters $\boldsymbol{\theta}$ along two random directions to generate a 3D loss landscape~\cite{li2018visualizing}.   Figure~\ref{figure1} (a) and (c) display the loss landscapes\footnote{Noted that the figure only shows partial loss surfaces since we only sample a 2D grid to plot the curves. The real loss landscapes should be much more complex.} of NGCF and LightGCN for  Amazon  dataset. First, despite using the same loss function, the loss surfaces of NGCF are relatively complicated compared to LightGCN, due to the use of non-linear  transformation in NGCF. Second, we notice that both NGCF and LightGCN may converge at   \textsl{sharp} minima, whose curvatures are unstable since  the loss values could change quickly around their neighborhood, leading to poor generalization~\cite{keskar2017on,li2018visualizing,foret2020sharpness,andriushchenko2022towards}.   Therefore, we argue that  the first-order optimizers (\textsl{e.g.,}  Adam) only seek the model parameters that minimizes the training error, but they dismiss the high-order information like the sharpness
of the loss landscape.  

In this work, we propose an effective training schema, called Graph-based Sharpness-Aware Minimization (gSAM), under the principle that the \textit{flatter}  minima has a better generalization ability than the  \textit{sharper} ones~\cite{keskar2017on,foret2020sharpness,andriushchenko2022towards}. gSAM explicitly
penalizes the sharp minima and biases the convergence to
a flat region  by forming a bi-level optimization: the outer problem conducts the standard model training while the inner problem helps the model jump out of the sharp minima. By doing so,  our  gSAM is able to produce   smoother loss surfaces as shown in Figure~\ref{figure1} (b) and (d). Finally, we extensively evaluate our proposed gSAM on several  benchmark datasets, obtaining favorable results compared with exiting GNNs.

\section{Related Work}
\subsection{Graph Neural Network}
 GNNs learn node representations by  aggregating  structural messages  from their neighbor, which have been successfully applied to collaborative filtering~\cite{ying2018graph,wang2019neural,he2020lightgcn,chen2022graph, chen2022adversarial}. Notably, GNN-based models, such as PinSage~\cite{ying2018graph}, NGCF~\cite{wang2019neural}, LightGCN~\cite{he2020lightgcn}, and MixGCF~\cite{huang2021mixgcf} have achieved superior performance in many applications.  However,  GNNs tend to yield inferior performance when the distributions of training and test data are not aligned well~\cite{wu2021self,yu2022graph}.  One prominent direction to improve generalization is contrastive learning~\cite{wu2021self,yu2022graph,lin2022improving} that applies different data augmentation methods to extract  features from unlabeled data.  Nevertheless, contrastive frameworks  (e.g., SGL~\cite{wu2021self}) often require a large batch size of comparing pairs. It has been observed that  a large batch size easily makes the models converge to bad or  sharp minima,  resulting in unexpected performance~\cite{keskar2017on}.

 \subsection{Sharpness-Aware Minimization}
Recent studies~\cite{Jiang2020Fantastic,keskar2017on,kwon2021asam,li2018visualizing,andriushchenko2022towards,kim2022fisher} have shown a strong correlation
between the sharpness of loss landscape and the generalization error on a large set of neural networks. In particular, Sharpness-Aware Minimization (SAM)~\cite{foret2020sharpness} aims to minimize both the loss value and loss sharpness within a maximization region around each parameter during training.   SAM has inspired several follow-up works. For example, ASAM~\cite{kwon2021asam} introduces adaptive sharpness with
a scale-invariant property that  adjusts the maximization region of weight space. 
FisherSAM~\cite{kim2022fisher} replaces SAM’s Euclidean balls with ellipsoids
induced by the Fisher information, which obtains more accurate  manifold structures. However, the use of a one-step gradient within SAMs is unstable with large variance. Recent efforts~\cite{liu2022random,demake} surprisingly find that adding random noise perturbations does not hurt the inner gradient ascent, implying the one-step gradient provides little gradient information. To overcome this limitation, we make a rigorous connection between SAM and bi-level optimization. We further put forward an implicit differentiation algorithm to consider the Hessian to achieve better generalization  for GNNs.

\section{PRELIMINARIES}

\subsection{Problem Setup}
	 In this work, we  focus on implicit recommendation, in which the behavior data involves a set of users $\mathcal{U}=\{u\}$, a set of items $\mathcal{I}=\{i\}$, and the observed user-item interactions $\mathcal{O}^+=\{y_{ui} | u \in \mathcal{U}, i \in \mathcal{I}\}$, where $y_{ui}$ denotes that user $u$ has interacted with item $i$ before. 
 
 One can view user-item interactions as a bipartite graph $\mathcal{G}=(\mathcal{V}, \mathcal{E})$~\cite{wang2019neural,he2020lightgcn}, where the nodes set $\mathcal{V}=\mathcal{U} \cup \mathcal{I}$ contains all users and items, and the edge set $\mathcal{E} = \mathcal{O}^+$ denotes the observed user-item links.  The goal of collaborative filtering is to recommend a ranked list of items  that are of interest to the user $u \in  \mathcal{U}$, in the same sense that performing link prediction on the bipartite graph $\mathcal{G}$.

\subsection{Graph Neural Network}
\subsubsection{\textbf{Message-passing Schema}}
  The core idea of GNNs is to  update the  representation of each node by aggregating messages from its neighbors, which can be expressed as:
\begin{equation}
\small
\mathbf{E}^{(l+1)} = f_{\text{agg}} (\mathbf{E}^{(l)}, \mathcal{G}),
\label{eq1}
\end{equation}
where $\mathbf{E}^{(l)}$ is the nodes' embeddings at the $l$-th layer, and $\mathbf{E}^{(0)}$ can be initialized via lookup tables;  $f_{\text{agg}}(\cdot)$ can be any differentiable aggregation functions~\cite{ying2018graph, wang2019neural,he2020lightgcn}. After $L$ layers, one may adopt a readout function to generate the final embeddings as:
\begin{equation}
\small
    \mathbf{e}_u = f_\text{readout}(\{\mathbf{e}^{(l)}_u, 0 \le l \le L\}),   \mathbf{e}_i = f_\text{readout}(\{\mathbf{e}^{(l)}_i, 0 \le l \le L\}),
\end{equation}
where the readout function $f_\text{readout}(\cdot)$ is usually permutation invariant like concatenation~\cite{wang2019neural} and weighted sum~\cite{he2020lightgcn}. Based on the final representations $(\mathbf{e}_u, \mathbf{e}_i)$, we can use inner product to predict how likely user $u$ would interact with item $i$ as: $\hat{y}_{ui} = \mathbf{e}_u^\top\mathbf{e}_i$.

\subsubsection{\textbf{Bayesian Personalized Ranking Loss}} 
One common objective function is the pairwise Bayesian Personalized Ranking (BPR) loss~\cite{rendle2009bpr}, which enforces the prediction of an observed interaction to be scored higher than its unobserved counterparts:
\begin{equation}
\small
\mathcal{L}_{\text{bpr}}(\bm{\theta}) = \sum_{(u, i, j) \in \mathcal{O}} - \ln \sigma(\hat{y}_{ui} - \hat{y}_{uj}),
\label{eq2}
\end{equation}
where $\mathcal{O}=\{(u,i,j)|(u,i)\in\mathcal{O}^{+},(u,j)\in\mathcal{O}^{-}\}$ is the training data, and $\mathcal{O}^{-}$ contains the unobserved user-item interactions; $\bm{\theta}$ denotes the model parameters in GNNs. However, obtaining good generalizations  for GNNs   is challenging.  We next revisit Sharpness-Aware Minimization~\cite{keskar2017on,kwon2021asam,li2018visualizing,andriushchenko2022towards}, which  exploits the relationship between sharpness/flatness of local minima and their
generalization.

\subsection{Sharpness-Aware Minimization}
\subsubsection{\textbf{PAC-Bayesian Bound}}
We begin by introducing the PAC-Bayesian theory~\cite{neyshabur2017exploring,huio}, which provides a foundation for deriving an upper bound of the generalization gap between the training and test error. The PAC-Bayesian theory is as follows:
\begin{theorem}[PAC-Bayesian Bound~\cite{neyshabur2017exploring,huio}]
Given a prior distribution $\mathcal{P}$ on the weight $\bm{\theta}$ of a neural network, any $\tau \in (0, 1]$, an expected error loss $\hat{\mathcal{L}}(\bm{\theta},\hat{\mathcal{D}})$ for a data distribution $\hat{\mathcal{D}}$, its empirical loss ${\mathcal{L}}(\bm{\theta}, {\mathcal{D}})$,  for any posterior distribution $\mathcal{Q}$ of the weight $\bm{\theta}$,  and let ${\mathcal{D}}$ drawn $m$ i.i.d. samples from $\hat{\mathcal{D}}$,  the following inequality holds with a probability at least $1-\tau$, 
\begin{equation}
\small
\mathbb{E}_{\bm{\theta} \in \mathcal{Q}}\hat{\mathcal{L}}(\bm{\theta},\hat{\mathcal{D}}) \leq  \mathbb{E}_{\bm{\theta} \in \mathcal{Q}}{\mathcal{L}}(\bm{\theta}, {\mathcal{D}}) + 4\sqrt{\frac{1}{m}\text{KL}(\mathcal{Q}||\mathcal{P}) + \log\frac{6m}{\tau}}.
    \label{pac}
\end{equation}
\end{theorem}

The goal of PAC-Bayesian learning is to optimize $\mathcal{Q}$ on the RHS of Eq.~(\ref{pac}) in order to obtain a tight upper bound on the test error $\mathbb{E}_{\bm{\theta} \in \mathcal{Q}}\hat{\mathcal{L}}(\bm{\theta},\hat{\mathcal{D}})$ (LHS). However, directly optimizing RHS of Eq.~(\ref{pac}) over $\mathcal{Q}$ is difficult because of the square root term.  

Instead of optimizing the RHS, Sharpness-Aware Minimization (SAM)~\cite{foret2020sharpness} replaces it with $\max_{\|\bm{\delta}\|_2 \leq \rho} {\mathcal{L}}(\bm{\theta}+\bm{\delta}, {\mathcal{D}})$ and uses $l_2$ norm on $\|\bm{\theta}\|_2$.

As such, SAM aims to minimize two losses~\cite{foret2020sharpness}: 1) the vanilla loss ${\mathcal{L}}(\bm{\theta}, {\mathcal{D}})$ that optimizes the model parameters, and 2) the loss associated
to the sharpness term ${\mathcal{R}}(\bm{\theta}, {\mathcal{D}})$ that maximizes change of the training loss within the local neighborhood:
\begin{equation}
\small
    \begin{aligned}
        \bm{\theta}^* = \arg\min_{\bm{\theta}} {\mathcal{L}}(\bm{\theta}, {\mathcal{D}}) + {\mathcal{R}}(\bm{\theta}, {\mathcal{D}}), \text{where}\\
         {\mathcal{R}}(\bm{\theta}, {\mathcal{D}}) =  \max_{\Vert \bm{\delta}\Vert _{2} \leq \rho } {\mathcal{L}}(\bm{\theta}+\bm{\delta}, {\mathcal{D}}) - {\mathcal{L}}(\bm{\theta}, {\mathcal{D}}),
    \end{aligned}
    \label{sam1}
\end{equation}
where $\rho$ is a constant radius. The Eq. (\ref{sam1}) is originally developed for i.i.d. data,  recent efforts~\cite{liao2020pac, ma2021subgroup}  show that the  PAC-Bayesian generalization
bound also holds for non-i.i.d. graph data.  Inspired by these findings, we can extend the  Eq. (\ref{sam1}) to our GNN-based collaborative filtering Eq. (\ref{eq2}), and rewrite the sharpness-aware  minimization problem as
 the following minimax
optimization:
\begin{equation}
\small
    \min_{\bm{\theta}} \max_{\Vert \bm{\delta}\Vert _{2} \leq \rho } {\mathcal{L}_{\text{bpr}}}(\bm{\theta}+\bm{\delta}).
    \label{eq6}
\end{equation}

Intuitively, Eq. (\ref{eq6}) minimizes the maximum loss around the neighborhood of $\bm{\theta}$. In this way, the maximum loss within the $\bm{\theta}$'s neighborhood area could be close to the loss of model parameters $\bm{\theta}$. Therefore, it expects to converge to a flatter minimum compared to minimizing the loss ${\mathcal{L}_{\text{bpr}}}(\bm{\theta})$ only. 

\subsubsection{\textbf{One-step Gradient}}
However, finding the exact solution for Eq. (\ref{eq6}) is NP-hard. SAM uses a one-step gradient ascent to approximate the
optimal solution. According to the first-order Taylor expansion, we can find the $\bm{\delta}^*$ as:
\begin{equation}
\small
\boldsymbol{\delta}^* \approx \underset{\|\boldsymbol{\delta}\|_2 \leq \rho}{\arg \max } ~ \mathcal{L}_{\text{bpr}}(\boldsymbol{\theta})+\boldsymbol{\delta} \cdot \nabla_{\boldsymbol{\theta}} \mathcal{L}_{\text{bpr}}(\boldsymbol{\theta})=\rho \cdot \frac{\nabla_{\boldsymbol{\theta}} \mathcal{L}_{\text{bpr}}(\boldsymbol{\theta})}{\left\|\nabla_{\boldsymbol{\theta}} \mathcal{L}_{\text{bpr}}(\boldsymbol{\theta})\right\|_2}.
\label{delta}
\end{equation}

With the determined $\boldsymbol{\delta}^*$, SAM  approximates $\nabla_{\boldsymbol{\theta}} \mathcal{L}_{\text{bpr}}(\boldsymbol{\theta}+\boldsymbol{\delta}^*)$ to $\nabla_{\boldsymbol{\theta}} \mathcal{L}_{\text{bpr}}(\boldsymbol{\theta})$ at $\boldsymbol{\theta} = \boldsymbol{\theta} + \boldsymbol{\delta}^*$, which  could be addressed by the gradient descent framework to obtain the $\boldsymbol{\theta}^*$. 

\begin{figure}[t]
\centering  \includegraphics[width=0.86\linewidth]{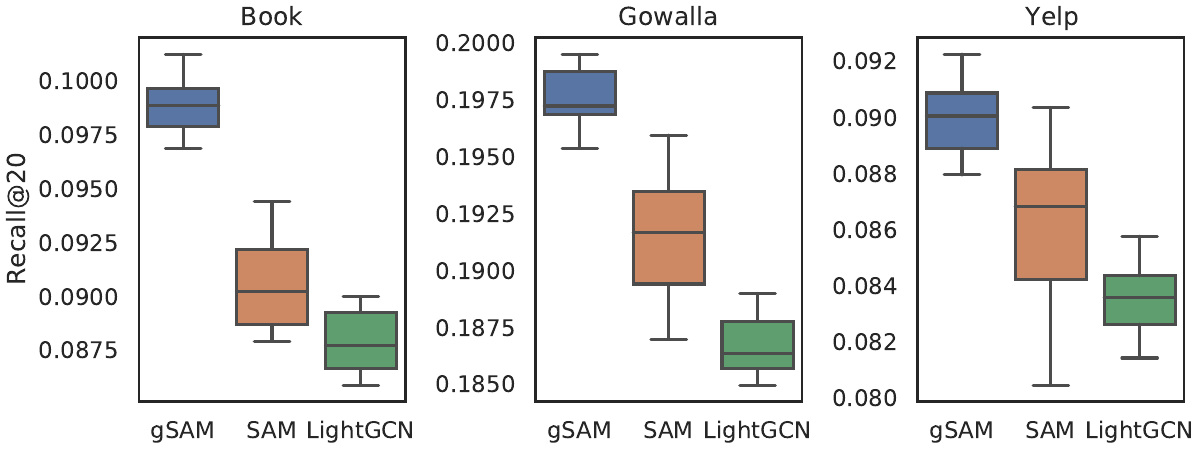}
\caption{Results of gSAM, SAM, and LightGCN for 30 runs. Compared to baselines, our proposed gSAM is more stable.}
\label{fig2}
\end{figure}

\subsubsection{\textbf{Limitation}}
The  one-step gradient algorithm works well for i.i.d. image data. However, we empirically find it lacks stability for non-i.i.d. graph data. Figure~\ref{fig2} shows the  results  of LightGCN, SAM, and our gSAM for $50$ runs on three benchmark datasets. Clearly, it can be seen that SAM (one-step gradient) has a large variance in performance. In some cases, e.g., Yelp dataset, the SAM  performs worse than  LightGCN. This implies that the one-step gradient algorithm is not sufficient to solve the minimax optimization.   In contrast, our gSAM consistently performs better than SAM and LightGCN. We next introduce our effective training schema in detail.

\section{The proposed gSAM}
Here we present our  gSAM by leveraging the implicit function theorem to understand the relationship between $\bm{\theta}$ and $\bm{\delta}$. In particular, we reformulate Eq. (\ref{eq6}) into a Bi-level Optimization problem~\cite{liu2021towards}:
\begin{equation}
\small
    \min_{\bm{\theta}} \quad  \mathcal{L}_{\text{out}}(\bm{\theta}, \bm{\delta}^*(\bm{\theta})), \quad
    \text{s.t.} \quad  \bm{\delta}^*(\bm{\theta}) = \argmin_{\|\bm{\delta}\|_2 \leq \rho } \mathcal{L}_{\text{in}} (\bm{\theta}, \bm{\delta}),
    \label{dd}
\end{equation}
where  $\mathcal{L}_{\text{out}}=-\mathcal{L}_{\text{in}}=\mathcal{L}_{\text{bpr}}$, and the implicit function $ \bm{\delta}^*(\bm{\theta})$ is the best-response of the model weights $\bm{\delta}$ to $\bm{\theta}$.  As such, we use the outer problem to conduct the standard model training, while the inner problem helps the model jump out of the sharp minima.

For the inner objective $\mathcal{L}_{\text{in}}$, we can use Projected Gradient Descent to update the $\bm{\delta}$, while for the outer objective  $\mathcal{L}_{\text{out}}$, we can decompose the hypergradient $\nabla_{\bm{\theta}}\mathcal{L}_{\text{out}}(\bm{\theta}, \bm{\delta}^*(\bm{\theta}))$ into:
\begin{equation}
\small
\nabla_{\bm{\theta}}\mathcal{L}_{\text{out}}(\bm{\theta}, \bm{\delta}^*(\bm{\theta})) = \frac{\partial \mathcal{L}_{\text{out}}(\bm{\theta}, \bm{\delta}^*(\bm{\theta}))}{\partial \bm{\theta}} +  \frac{\partial \mathcal{L}_{\text{out}}(\bm{\theta}, \bm{\delta}^*(\bm{\theta}))}{\partial \bm{\delta}^*(\bm{\theta})} \times \frac{\partial \bm{\delta}^*(\bm{\theta})}{\partial \bm{\theta} },
\end{equation}
where the first term denotes the direct gradient that is easy to compute, while the second term is the indirect gradient where we must compute  the Jacobian $\frac{\partial \bm{\delta}^*(\bm{\theta})}{\partial \bm{\theta} }$. Inspired by  Cauchy Implicit Function theorem~\cite{lorraine2020optimizing}, we can estimate the Jacobian as:
\begin{equation}
\small
\left.\frac{\partial \bm{\delta}^*(\bm{\theta})}{\partial \bm{\theta} }\right|_{\bm{\theta'}} = \left. -\left[\frac{\partial^2 \mathcal{L}_{\text{in}} (\bm{\theta}, \bm{\delta})}{\partial \bm{\delta} \partial \bm{\delta}^T}\right]^{-1} \times   \frac{\partial^2 \mathcal{L}_{\text{in}} (\bm{\theta}, \bm{\delta})}{\partial \bm{\delta} \partial \bm{\theta}^T}    \right|_{\bm{\delta}^*(\bm{\theta'}), \bm{\theta'}}
\end{equation}
Moreover, we can efficiently compute the inverse Hessian using Neumann series~\cite{liao2018reviving}:
\begin{equation}
\small
 \left[\frac{\partial^2 \mathcal{L}_{\text{in}} (\bm{\theta}, \bm{\delta})}{\partial \bm{\delta} \partial \bm{\delta}^T}\right]^{-1} = \lim_{i \to \infty} \sum_{j=0}^{i}\left [ \mathbf{I} - \frac{\partial^2 \mathcal{L}_{\text{in}} (\bm{\theta}, \bm{\delta})}{\partial \bm{\delta} \partial \bm{\delta}^T}  \right]^j,
\end{equation}
where $\mathbf{I}$ is an identity matrix. Usually, only the first $J$ terms of the Neumann series are  enough for approximation. As such, the final hypergradient $\nabla_{\bm{\theta}}\mathcal{L}_{\text{out}}(\bm{\theta}, \bm{\delta}^*(\bm{\theta}))$ can be computed as:
\begin{equation}
\small
\nabla_{\bm{\theta}}\mathcal{L}_{\text{out}} \approx \frac{\partial \mathcal{L}_{\text{out}}}{\partial \bm{\theta}} -  \frac{\partial \mathcal{L}_{\text{out}}}{\partial \bm{\delta}} \times \sum_{j=0}^{J}\left [ \mathbf{I} - \frac{\partial^2 \mathcal{L}_{\text{in}} }{\partial \bm{\delta} \partial \bm{\delta}^T}  \right]^j \times  \frac{\partial^2 \mathcal{L}_{\text{in}}}{\partial \bm{\delta} \partial \bm{\theta}^T}.
\label{haha}
\end{equation}

To this end, we can summarize the proposed gSAM as:
\begin{enumerate}
    \item Update  $\bm{\delta} \gets \mathcal{P}_\rho [\bm{\delta}- \eta_1 \nabla_{\bm{\delta}}\mathcal{L}_{\text{in}}]$, where $\mathcal{P}_\rho[\mathbf{x}] = \rho \frac{\mathbf{x}}{\max\{\rho, \|\mathbf{x}\|_2 \}} $ denotes the projection onto the $l_2$ ball of radius $\rho$.
    \item Update  model parameters  $\bm{\theta} \gets \bm{\theta}- \eta_2 \nabla_{\bm{\theta}}\mathcal{L}_{\text{out}}$ via Eq. (\ref{haha}).
\end{enumerate}
Unlike  the one-step gradient algorithm, our hypergradient algorithm can adjust each dimension of the gradient by Hessian, which is more tolerant to the change of the loss curvature. More importantly, our implicit hypergradient can be implemented in a memory-efficient manner compared to the explicit way of solving bi-level optimization~\cite{liu2021towards}, which is scalable to millions of parameters~\cite{lorraine2020optimizing}.

\begin{table}
\small
  \centering
  \caption{Statistics of three benchmark 
 datasets.}
  \label{tab3}
  \begin{tabular}{cccccc}
    \toprule
    Dataset & \#user & \#item &  \#inter  &  inter/user & density\\
    \midrule
    Book & 52.6k & 91.6k & 2984.1k & 56.7 & 0.06\%\\
    Gowalla  & 29.9k & 41.0k &  1027.4k & 56.7 & 0.06\%\\
    Yelp & 31.7k & 38.0k & 1561.4k & 49.3 & 0.13\%\\
  \bottomrule
  \vspace{-6mm}
\end{tabular}
\end{table}

\section{Experiments}
\subsection{Experimental Settings}
\subsubsection*{{\textbf{Dataset}.}} We conduct experiments on three  datasets\footnote{https://github.com/kuandeng/LightGCN/tree/master/Data}: \textsl{Amazon-Book}, \textsl{Gowalla}, and \textsl{Yelp-2018}. The statistics of the datasets are summarized in Table~\ref{tab3}. For each dataset,  we randomly split each user’s historical interactions into
training/validation/test sets with the ratio  8:1:1. Also we adopt two common used Top-$k$ metrics: Recall@$k$ and NDCG@$k$ ( $k=20$ by default) with  the all-ranking protocol~\cite{he2020lightgcn}.

\subsubsection*{{\textbf{Baselines}.}} Our gSAM is fully compatible with existing GNN-based recommenders to obtain better generalization. We choose the following baselines:  1) \textbf{NGCF}~\cite{wang2019neural}, which applies the message-passing scheme   to exploit the high-order neighbors' information; 2) \textbf{LightGCN}~\cite{he2020lightgcn}, which omits the non-linear transformation   to obtain node representations; 3) \textbf{MixGCF}~\cite{huang2021mixgcf}, which uses the mix-up strategy to generate hard negative samples; 4) \textbf{SimGCL}~\cite{yu2022graph} is  a   contrastive framework that  adds uniform noise  to  the  representations. 
	
\subsubsection*{{\textbf{Parameter Settings}.}} For all baselines, the size of user/item representation is searched among $\{32, 64, 128, 256\}$. For NGCF, LightGCN, MixGCF, and SimGCL, their hyperparameters are initialized the same as their original settings, and are then carefully tuned to achieve optimal performance. For gSAMs, we choose the same hyperparameters as their backbones,  such as batch size, stopping criteria, learning rate, \textsl{etc}. For the radius of the neighbor ball $\rho$ in Eq. (\ref{dd}),   we vary $\rho$ within $\{ 0.01, 0.05, 0.1, 0.5, 1.0\}$.

	\subsection{Experimental Results}
	\subsubsection{\textbf{Overall Performance}}
	The results of different models in terms of Recall$@20$ and NDCG$@20$  are summarized in Table~\ref{tab4}. From the experimental results, we mainly have the following observations. First, MixGCF and SimGCL perform better than NGCF and LightGCN. MixGCF synthesizes  hard negative samples using mix-up data augmentation, while SimGCL adopts random noises data augmentation. This indicates that the GNN models generally get benefits from data augmentation during the training. 
	Second, NGCF+gSAM and LightGCN+gSAM yield better performance than their backbones for all datasets. For example, by comparing the LightGCN and LightGCN+gSAM, LightGCN+gSAM has on average $9.34\%$ improvement with respect to Recall$@20$ and over $8.85\%$ improvements in terms of NDCG$@20$. This verifies the necessity of  explicitly smoothing the loss geometry during model training. 
	Third, gSAM can further improve the accuracy of MixGCF and SimGCL  with a large margin, which  shows the potential benefit of integrating  sharpness-aware minimization and data augmentation.

\begin{table}[]
\centering
\caption{The performance of gSAM with different backbones.}
\scalebox{0.70}{\begin{tabular}{c|cc|cc|cc}
\toprule[1.2pt]
         & \multicolumn{2}{c|}{Book}                                                                                                 & \multicolumn{2}{c|}{Gowalla}                                                                                              & \multicolumn{2}{c}{Yelp}                                                                                                  \\
Method   & recall                                                      & ndcg                                                        & recall                                                      & ndcg                                                        & recall                                                      & ndcg                                                        \\\midrule
NGCF     & 0.0759                                                      & 0.0466                                                      & 0.1373                                                      & 0.0810                                                      & 0.0716                                                      & 0.0440                                                      \\
\rowcolor[HTML]{EFEFEF} 
+gSAM    & \begin{tabular}[c]{@{}c@{}}0.0822\\ (+8.30\%)\end{tabular}  & \begin{tabular}[c]{@{}c@{}}0.0504\\ (+8.15\%)\end{tabular}  & \begin{tabular}[c]{@{}c@{}}0.1551\\ (+12.96\%)\end{tabular} & \begin{tabular}[c]{@{}c@{}}0.0954\\ (+17.77\%)\end{tabular} & \begin{tabular}[c]{@{}c@{}}0.0805\\ (+12.43\%)\end{tabular} & \begin{tabular}[c]{@{}c@{}}0.0488\\ (+10.91\%)\end{tabular} \\\midrule
LightGCN & 0.0875                                                      & 0.0576                                                      & 0.1865                                                      & 0.1086                                                      & 0.0833                                                      & 0.0514                                                      \\
\rowcolor[HTML]{EFEFEF} 
+gSAM    & \begin{tabular}[c]{@{}c@{}}0.0993\\ (+13.48\%)\end{tabular} & \begin{tabular}[c]{@{}c@{}}0.0641\\ (+11.28\%)\end{tabular} & \begin{tabular}[c]{@{}c@{}}0.1977\\ (+6.01\%)\end{tabular}  & \begin{tabular}[c]{@{}c@{}}0.1163\\ (+7.09\%)\end{tabular}  & \begin{tabular}[c]{@{}c@{}}0.0904\\ (+8.52\%)\end{tabular}  & \begin{tabular}[c]{@{}c@{}}0.0556\\ (+8.17\%)\end{tabular}  \\\midrule
MixGCF   & 0.0922                                                      & 0.0601                                                      & 0.2011                                                      & 0.1221                                                      & 0.0889                                                      & 0.0546                                                      \\
\rowcolor[HTML]{EFEFEF} 
+gSAM    & \begin{tabular}[c]{@{}c@{}}0.0998\\ (+8.36\%)\end{tabular}  & \begin{tabular}[c]{@{}c@{}}0.0655\\ (+9.03\%)\end{tabular}  & \begin{tabular}[c]{@{}c@{}}0.2162\\ (+7.54\%)\end{tabular}  & \begin{tabular}[c]{@{}c@{}}0.1335\\ (+9.32\%)\end{tabular}  & \begin{tabular}[c]{@{}c@{}}0.0962\\ (+8.26\%)\end{tabular}  & \begin{tabular}[c]{@{}c@{}}0.0597\\ (+9.33\%)\end{tabular}  \\ \midrule
SimGCL   & 0.0941                                                      & 0.0642                                                      & 0.1986                                                      & 0.1189                                                      & 0.0937                                                      & 0.0571                                                      \\
\rowcolor[HTML]{EFEFEF} 
+gSAM    & \begin{tabular}[c]{@{}c@{}}0.1016\\ (+7.97\%)\end{tabular}  & \begin{tabular}[c]{@{}c@{}}0.0721\\ (+12.31\%)\end{tabular} & \begin{tabular}[c]{@{}c@{}}0.2152\\ (+8.36\%)\end{tabular}  & \begin{tabular}[c]{@{}c@{}}0.1287\\ (+8.24\%)\end{tabular}  & \begin{tabular}[c]{@{}c@{}}0.1024\\ (+9.28\%)\end{tabular}  & \begin{tabular}[c]{@{}c@{}}0.0629\\ (+10.16\%)\end{tabular} \\ \toprule[1.2pt]
\end{tabular}}
\label{tab4}
\end{table}

\subsubsection{\textbf{Further Probe}}
\begin{figure}[t]
\centering  \includegraphics[width=0.80\linewidth]{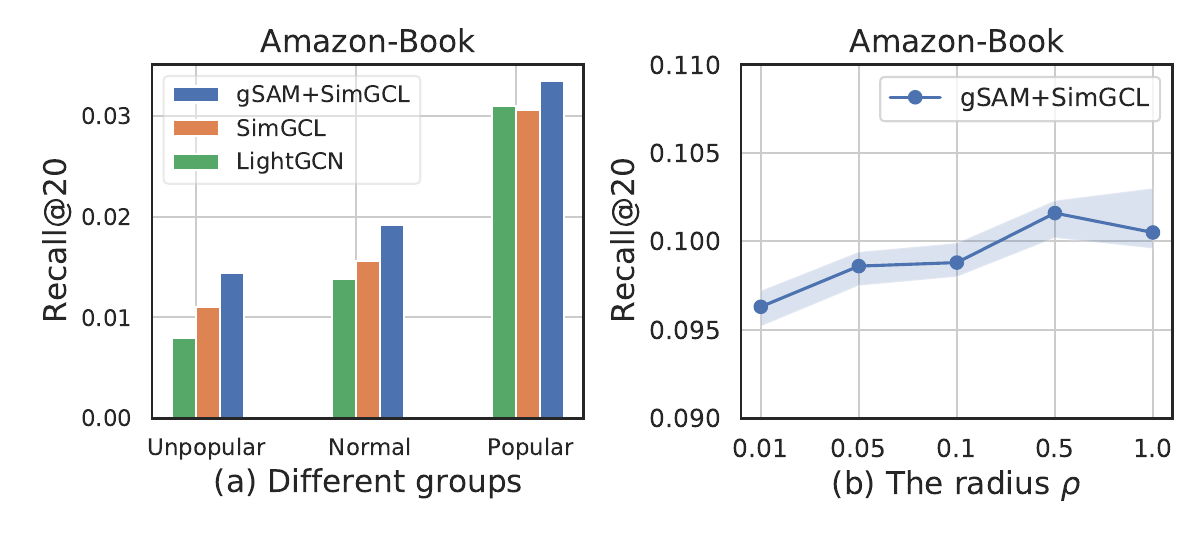}
\caption{(a) Performance comparison for different item groups. (b) Influence of the radius $\rho$.}
\label{fig3}
\vspace{-3mm}
\end{figure}

 GNNs are known to be  biased towards high-degree items, neglecting the impact of low-degree items. Here we investigate the generalization ability of our gSAM on Book dataset (the other two have similar results and are omitted here). Following \cite{yu2022graph}, we divide the test set into three subsets in proportion to the popularity of items: 'Unpopular', 'Normal', and 'Popular'. From Figure \ref{fig3}(a), we can find
that the performance of gSAM is consistently better than LightGCN and SimGCL. This implies that gSAM generally has better generalization as it can still
perform high-quality recommendation with sparse data.

 Our gSAM has an additional hyperparameter: the  radius $\rho$ in Eq. (\ref{dd}).  To analyze the influence of
$\rho$, we vary  $\rho$ in the range of $0.01$ to $1.0$ and report the experimental results in
Figure \ref{fig3}(b). We observe that our gSAM is stable with respect to  $\rho$. Specifically, when $\rho$  is set to around $0.5$, we obtain the best performance on Amazon dataset.

\section{conclusion}
Training GNN-based recommenders can easily fall into sharp minima, which may lead to poor generalization.  To address this issue, we present a novel training framework, call gSAM, to smooth out the loss landscapes  during the training of GNNs. The core idea behind gSAM is to explicitly penalize the sharp minima and guide the convergence to a flatter region by solving
a bi-level optimization. Extensive experimental results demonstrate the positive impact of our gSAM for personalized ranking with better generalization.

\bibliographystyle{ACM-Reference-Format}
\balance
\bibliography{sample-base}

\end{document}